\documentclass[numsec,webpdf,modern,medium]{oup-authoring-template}%

\usepackage{makecell} 
\usepackage{array} 

\graphicspath{{Fig/}}

\def \makeL
{\hskip-0.25em'}

\setcitestyle{round}
\begin{document}

\journaltitle{Submitted to the Digital Scholarship in the Humanities}
\DOI{DOI HERE}
\copyrightyear{January 2025}
\pubyear{2025}
\access{Advance Access Publication Date: Day Month Year}
\appnotes{Paper}

\firstpage{1}


\title[Iconicity in Large Language Models]{Iconicity in Large Language Models}
\author[1]{Anna Marklová\ORCID{0000-0003-3392-1028}}
\author[1,$\ast$]{Jiří Milička\ORCID{0000-0001-8605-1199}}
\author[2]{Leonid Ryvkin\ORCID{0000-0003-1634-5372}}
\author[3]{L\makeL udmila Lacková Bennet\ORCID{0000-0001-9852-4280}}
\author[3]{Libuše Kormaníková\ORCID{0000-0001-8919-6748}}

\authormark{Marklová, Milička et al.}

\address[1]{\orgdiv{Department of Linguistics}, \orgname{Charles University}, \orgaddress{\street{Street}, \postcode{Postcode}, \state{State}, \country{Country}}}
\address[2]{\orgdiv{Department}, \orgname{Claude Bernard University Lyon}, \orgaddress{\street{Street}, \postcode{Postcode}, \state{State}, \country{Country}}}
\address[3]{\orgdiv{Department of General Linguistics}
, \orgname{Palacký University Olomouc}, \orgaddress{\street{Street}, \postcode{Postcode}, \state{State}, \country{Country}}}

\corresp[$\ast$]{Corresponding author. \href{email: jiri@milicka.cz}{jiri@milicka.cz}}

\received{Date}{0}{2025}
\revised{Date}{0}{Year}
\accepted{Date}{0}{Year}



\abstract{Lexical iconicity, a direct relation between a word's meaning and its form, is an important aspect of every natural language, most commonly manifesting through sound-meaning associations. Since Large language models' (LLMs') access to both meaning and sound of text is only mediated (meaning through textual context, sound through written representation, further complicated by tokenization), we might expect that the encoding of iconicity in LLMs would be either insufficient or significantly different from human processing. This study addresses this hypothesis by having GPT-4 generate highly iconic pseudowords in artificial languages. To verify that these words actually carry iconicity, we had their meanings guessed by Czech and German participants (n=672) and subsequently by LLM-based participants (generated by GPT-4 and Claude 3.5 Sonnet). The results revealed that humans can guess the meanings of pseudowords in the generated iconic language more accurately than words in distant natural languages and that LLM-based participants are even more successful than humans in this task. This core finding is accompanied by several additional analyses concerning the universality of the generated language and the cues that both human and LLM-based participants utilize.}



\maketitle

\section{Introduction}

Lexical iconicity is (in simplified terms) a direct relation between word's meaning and its form, and although it was marginalized by many 20th-century linguists starting with de Saussure, it remains an important aspect of language. It plays a role in acquisition \citep{perry2018iconicity} and is a consequence of our embodiment \citep{vigliocco2014language}. It will therefore be interesting to examine how large language models, which by default lack embodiment and acquire language in a technically different way than humans \citep{huebner-etal-2021-babyberta}, process iconicity. Surprisingly, there are very few studies investigating iconicity in large language models (with exceptions such as \citet{trott2024can} and \citet{loakman2024earseyeshearsound}). This paper aims to fill this gap.

The very definition of iconicity suggests that large language models should struggle with it, as their processing of both semantic and phonetic aspects of language is indirect:
\begin{enumerate}
\item Regarding lexical semantics, the vector represen\-tation of semantics in language models based on embeddings processed by transformers \citep{bahdanau2016neuralmachinetranslationjointly, vaswani2017attention} does not describe a direct relationship between the signifier and the signified, as de Saussure conceived it, but rather encodes relationships between symbols through their distributional patterns. While this meaning-through-usage approach is not alien to humans, as argued by later Wittgenstein \citep{giesewetter2014meaning} and empirically demonstrated by corpus linguists \citep{sinclair1991corpus}, humans still possess direct experiential knowledge that is not mediated through text.
\item The phonetic aspect of language is likewise only mediated, and language models' training data do not contain audio recordings.
The current architecture, which relies on BPE tokenization \citep{Gage1994BPE}, is particularly problematic in this regard \citep{bostrom2020byte, yehudai2024transformerscountn}. Even if we somehow incorporated wave form representations into the models, it would still be merely a symbolic representation of those waves, failing to evoke the physical reactions that humans experience when perceiving iconic words \citep{peeters2016processing}, not to mention that iconicity in its general definition concerns not only sound but also manner of articulation, position of speech organs (for instance, lips are rounded for [o], which a model based entity cannot experience, as it has no actual lips).
\end{enumerate}

LLMs thus access iconicity only through academic texts about iconicity and etymology, non-academic discourse on these topics, literary analyses, and literature itself, namely children 
literature and poetry that relies heavily on onomatopoia and iconicity in general. However, this does not necessarily prevent LLMs from working with iconicity. In fact, the perception of the entire physical world is mediated through human-generated texts and thus challenging for LLMs. Nevertheless, LLMs can simulate various aspects of physical world \citep{wang2023newton} and human cognition quite effectively, not only basic reasoning \citep{han2024inductive}, but also more subtle aspects of human cognition such as various cognitive biases \citep{suri2024large}, human-like learning \citep{wang2024llmsimaginariumtoollearning}, or theory of mind \citep{milicka2024large}. When we prompt an anthropomorphic entity, that anthropomorphic entity behaves anthropomorphically, as the given architecture functions as a universal simulator \citep{shanahan2023role}. Even models that have undergone Reinforcement Learning from Human Feedback (RLHF) instruction tuning can simulate human personas in various life situations and from different social groups \citep{deWinter2024questionnaires}.

There are thus arguments both for why LLMs would fail to process iconicity (because it is at the intersection of two planes of language that is hard to access for them and it is a low-level feature of language that is hard to simply pattern match) and arguments for them being able to do it (because LLMs emergently capture hard to pattern match aspects of human cognition and behavior). There is no other way how to decide between those two predictions than to look at empirical data.


Hence, to empirically investigate these competing theoretical predictions, we designed a two-phase experimental study. The first phase examines whether LLMs (represented by GPT-4) can generate pseudowords with high iconicity. By prompting GPT-4 to create a lexicon for an artificial language with strong sound-symbolic properties, and then having human participants guess the meanings of these generated words, we can assess whether the model can successfully encode iconic relationships that are interpretable by humans. These results will be benchmarked against comparable findings from natural languages documented in \citet{NasClanekVeSpeechi}, a study that has been done using the same methodology and performed on the same set of lexical meanings. In the second phase we use LLM-based participants to guess the meanings of these iconic pseudowords, allowing us to examine whether models can not only generate but also perceive and interpret sound-symbolic relationships. 

All data and scripts used for word generation and analysis, along with the \emph{Protocol} that describes details of all experiments and their results, are openly available at \url{https://osf.io/ywjrk/}.

\section{Motivation and aims}

A common definition of iconicity is that it is a property which holds when signs resemble their referents \citep{Vinson2021}. Iconicity can also be characterized by its role in human cognition and communication. \citet{Motamedietal2019} offer a functional definition that states that iconicity is a feature of a signal that enables its meaning to be predicted from its form. Iconicity, together with its complementary force, arbitrariness \citep{Saussure1983}, are widely considered to be present across all spoken and signed languages \citep{Dingemanse2016, Perniss2014}. The question is, can iconicity be created artificially? 

There were attempts to create artificial languages that would be universally well learnt and understood. Esperanto \citep{Zamenhof1912} is the most well-known of such attempts. Its vocabulary is largely based on European languages, grammar is similar to Indo-European languages, but often uses agglutination \citep{Kramer2023}. The high systematicity of Esperanto leads to its higher intelligibility (for example, negation of positive words is systematically created by the prefix `mal-' , such as `sana' \emph{healthy} - `malsana' \emph{sick}). \footnote{We will note here also attempts to create a pictorial orthographical system, such as IconText \citep{Beardon1995}, PECS \citep{Frost2002}, or Pictoperanto \citep{Gros2011} \cite[see][for an overview]{Kramer2023}}. Unlike Esperanto vocabulary, the LLM in our study was prompted to create vocabulary of a brand new language, and it was encouraged to use iconicity as a principle. We tried to avoid similarities with other languages by a curated prompt. 

The example of the most pronounced iconicity are onomatopoeia. These words phonologically imitate sounds in the real world \citep{Vinson2021}, and because of this feature, these words are easier to learn and remember than other ones \citep{perniss-thompson-vigliocco-2010, Lupyan2018, Roders2022}. Next to onomatopoeia, iconicity manifested through sound symbolic associations have been described. This iconicity emerges from various features of phonemes and their associations with characteristics of referents --- e.g., voiced stops as /b/, /d/, /g/, are associated with round shapes, while voiceless stops such as /p/, /t/, /k/, with spiky shapes \citep{kohler-1929, Sidhu2018}.

Iconicity as a fundamental principle in language has been explained by analogies between cultural evolution and natural selection: linguistic forms that are easier to process and remember will be used more often and spread faster than more difficult alternatives \citep{Vinson2021, Christiansen2008}. 

Empirical approaches to iconicity are of various kinds \cite[see][for an overview]{Motamedietal2019}. Comprehension experiments employing forced-choice methodology are widespread (e.g., ``does `bouba' mean round or spiky?") \cite[e.g.,]{Lockwood2016, Klima1979, Dingemanse2016}. Another common method are iconicity ratings. Speakers are asked how well a word resembles its referent, e.g. \citet{Perry2015, Winter2017, winter2024IconicityRatings14000}. Rarer are studies focusing on production, e.g. \cite{Nygaard2009, Perlman2015}. Data-driven approaches often focus on a more vague concept of iconicity. By mapping systematical statistical regularities across languages, these approaches often identify broader non-arbitrariness, such as systematicity \citep{Blasi2016, Haynie2014, Urban2011, Wichmann2010}. 

Recent studies have begun to explore the relationship between iconicity and AI. These studies are scarce and they usually focus rather on multimodality and vision language models (VLM) than LLMs. E.g., the traditional shape symbolism kiki-bouba paradigm \citep{ramachandran-hubbard-2001} has been explored in vision models (CLIP, Stable Diffusion) by \citet{alper2024kikiboubasoundsymbolism} and a certain level of sound symbolism was identified. \citet{loakman2024earseyeshearsound} explored LLMs' and VLMs' abilities to demonstrate sound symbolism in multiple experiments. They replicated the 
kiki-bouba experiment \citep{ramachandran-hubbard-2001} and discovered that overall, VLMs show rather low agreement with human assessments. Similarly, magnitude symbolism (based on the
non-arbitrary relationship between vowels and physical size, where e.g., /i/ in `mil' is associated with small entities, and a in `mal' is associated with larger entities) brought mixed results, with some models and conditions having low agreement with humans, and some reaching a higher agreement (GPT-4 achieving the most consistent agreement overall). Lastly, they let LLMs to rate iconicity, replicating \citet{winter2024IconicityRatings14000} research on 14k+ English words. They discovered a low agreement between human judgements and LLM ratings overall, but they made connection with the correlation and model size. The largest models had generally higher correlation with human judgements than the smaller ones. GPT-4 presented the strongest Spearman correlation of 0.537. They noted that it is interesting that they replicated this research after \citet{trott2024LLMs}, who used older GPT-4 model but achieved a higher Spearman correlation (0.575) than them, even though they performed it on a newer model. \citet{loakman2024earseyeshearsound} hypothesize about the sources of emergence of iconicity in LLMs/VLMs: the regular orthography of languages conveys associations between grapheme sequences and semantics, that can help learning of sound-based phenomena. Additionally, multimodal systems might contain the connection of, for example, high front vowels and pictures of small entities in their training data.

The current study follows up on our previous research on intelligibility of distant languages. It uses the same design of the experiment and focuses on the same measurements while tracking cues for intelligibility. In our original study \citep{NasClanekVeSpeechi}, we explored the ability of Czech speakers to guess meanings of words in four languages they do not speak, specifically, Turkish, Latvian, Hindi, and Japanese. These languages are of various affinity with Czech. Latvian as a Baltic language has many cognates with Czech, Hindi has the same Indo-European root but is in large geographical distance, Turkish has no common root with Czech but is geographically fairly close, and Japanese is geographically distant and shares no developmental root with Czech. In the study, we semi-randomly chose 64 words of different frequencies and obtained their forms in all five languages. These words were not sound-symbolic. In the task, a Czech speaker saw a pair of Czech words and their translations and was asked to pair them with the correct translations in an unknown language. In this study, we discovered that while in the closest language, Latvian, the intelligibility is the highest (80.3\% (CI 78.8--81.7\%), even in distant languages, the correctness rate is close to 60\% (resp., Hindi 60.9\% (CI 59.6--62.3\%), Turkish 57.3\% (CI 56.5--58.0\%), Japanese 60.0\% (CI 58.8--61.2\%). We tracked various characteristics to search for the features of the words that might have been used by the speakers to help them guess the meanings of unknown words. We discovered that word length agreement, vector and edit phonological similarities, and presence of cognates had positive influence on the chance that the word will be paired correctly. However, when asked, participants did not report that they would consciously look for such cues.

In the current paper, we ask if LLMs can invent words containing cues about their meanings in their forms. We will test if they were successful by presenting these words to speakers of Czech and German in the same design as in \citep{NasClanekVeSpeechi}. By prompting the model both in Czech and German, we will additionally test if the model created the words (from now on called ``pseudowords'') in some language-specific way, tailoring them to be intelligible to Czech (or German, respectively), or if the model employ some ``universal iconicity'', meaning, the intelligibility of the created pseudowords will not depend on the prompt language and the mother-tongue of the participants. Moreover, we explore if LLMs themselves are able to encode the meanings of pseudowords they created. For that, we will present the pairs of pseudowords and their possible meanings to GPT-4 and Claude 3.5 Sonnet.

This study aims to uncover (1) if LLMs are able to extract meta-linguistic information about iconicity and non-arbitrariness from their training data; (2) if they tailor the output according to the language of the prompt, creating a language that is most intelligible for the particular user, or if they create ``universally intelligible" language, (3) if the same cues influencing the intelligibility of the natural unknown words (i.e., length agreement, phonological similarity) will be used by speakers to guess meanings of the pseudowords, and (4) if LLMs themselves will be able to guess meanings of generated pseudowords.

\section{Methodology}
Our experimental design follows a classic approach used in other empirical studies on iconicity (e.g., \cite{kunihira-1971}). Each participant is presented with four stimuli: two of which are meaningful (e.g., words in a familiar language or images) and two that are primarily formal for the participant (e.g., pseudo-words or words in an unfamiliar language, either written or spoken). Participants are asked to match the two meaningful stimuli with the two formal stimuli.

The protocol about the exact run of the experiment, allowing replication, together with the full dataset, detailed demographic information about participants, and scripts of the experiment, can be found in Supplementary information.  

\subsection{Materials}
To maintain comparability with previous studies, we used the same dataset of words as in the article \emph{Sources of Intelligibility of Distant Languages: An Empirical Study}. Originally, we randomly selected 68 words from a frequency wordlist derived from the Hindi corpus HindEnCorp \cite{diatka2017effect,bojar2014hindencorp}. The selection adhered to specific criteria:

\begin{itemize}
    \item 22 words from the lower-frequency level (with 1–13 occurrences in the corpus);
    \item 23 words from the middle-frequency level (with 14–200 occurrences in the corpus);
    \item 23 words from the upper-frequency level (with over 200 occurrences in the corpus).
\end{itemize}

This stratified sampling approach ensures that the word set represents various frequency levels in the language. The full list with all words in English, Czech, and German, together with their pseudowords, can be found in the attachment. These words were translated to Czech and German. By using the same word set as previous studies, we maintain consistency and enable meaningful comparisons of results across different studies of the same research project.

Czech and German were chosen for their specific linguistic properties and the convenience of conducting research in a familiar environment. We selected Czech as the primary language because the Czech graphemic system is fully phonological, making it a particularly suitable choice for this study, since the participants can easily create a consistent mental representation of the written word in an unknown language when presented in Czech transcription. To increase the generalizability of our findings, we repeated the experiments with German speakers, as they too can readily derive the sound form of a word from its written form. Additionally, both languages represent smaller languages inside the GPT-4 training data. We consider it crucial to focus attention on languages other than English when exploring the potential of LLMs, since most LLM users do not have English as their mother-tongue. 

\subsubsection{Generating pseudowords}
The 68 words were translated into an artificial language created by \emph{GPT-4} by OpenAI. The exact scripts of the prompts and the answers given by the model can be found in Supplementary information. We generated two datasets of pseudowords: one using Czech as the language of the prompts, leading to the list of pseudowords we will call here \textit{Czech-prompted pseudowords}; second using German as the prompting language, leading to the list of \textit{German-prompted pseudowords}. The German and Czech prompt with English translations and an example of an answer can be seen in Table \ref{tab:Prompts}.

The prompt was refined after testing several versions, including one that did not specify the language features (i.e., asking to generate words from an imaginary language) and one that requested a fantasy language. The final prompt worked best for the experiment's purposes: Including the detail that the language is made-up ensured that the generated lexemes were not translations from a foreign language. Additionally, specifying a geographic area prevented the lexemes from closely resembling words from the original prompt's language, a problem that occurred when we either did not specify the area or instructed that the language was from a fantasy world (e.g., the Czech word \textit{zvíře} `animal' was translated as \textit{zvířeto}).

\begin{table*}[t]
\caption{German Prompting Session for Tubar Language (GPT-4).}\label{tab:tubar_prompting}
\tabcolsep=0pt
\begin{tabular*}{\textwidth}{@{\extracolsep{\fill}}p{0.45\textwidth}p{0.45\textwidth}@{\extracolsep{\fill}}}
\toprule%
\textbf{German} & \textbf{English} \\
\midrule
\textbf{System prompt:} & \textbf{System prompt:} \\
Lass uns ein Spiel spielen. Wir werden über den fiktiven Tubar-Stamm sprechen. Dieser südamerikanische Stamm hat seine eigene Sprache. Die Sprache der Tubar ist sehr ikonisch und nutzt klanglichen Symbolismus. Allein durch den Klang können Sie erahnen, was ein Wort bedeutet. Stellen Sie sich vor, Sie sind eine Expertin für diese Sprache. Wir geben Ihnen jetzt einige Bedeutungen. Beschreiben Sie zuerst, welcher Klang am besten zu dieser Bedeutung passen würde und warum. Dann denken Sie sich ein Wort in der Sprache des Tubar-Stammes mit dieser Bedeutung aus: &
Let us play a game. We will talk about the fictional Tubar tribe. This South American tribe has its own language. The Tubar language is very iconic and uses sound symbolism. Just from the sound, you can guess what a word means. Imagine you are an expert in this language. We will now give you some meanings. First, describe which sound would best match this meaning and why. Then, create a word in the Tubar language with this meaning: \\
\midrule
\textbf{User prompt:} & \textbf{User prompt:} \\
Bewegung & Movement \\
\midrule
\textbf{Response:} & \textbf{Response:} \\
Bewegung ist etwas Fließendes, Dynamisches und Kontinuierliches. In der Tubar-Sprache könnte dies durch lange, gleitende Vokale und fließende Konsonanten dargestellt werden, um die Idee von Bewegung zu vermitteln. \newline
Ein passendes Wort in der Tubar-Sprache könnte ``Looana" sein. ``Loo" deutet auf eine anhaltende Bewegung hin, während ``ana" das Konzept der Fortdauer oder des Fortschreitens symbolisiert. &
Movement is something flowing, dynamic, and continuous. In the Tubar language, this could be represented by long, gliding vowels and flowing consonants to convey the idea of movement. \newline
A suitable word in the Tubar language could be ``Looana." ``Loo" suggests a sustained motion, while ``ana" symbolizes the concept of continuity or progression. \\
\botrule
\end{tabular*}
\label{tab:Prompts}
\end{table*}

The results were obtained by gpt-4-0613 model, sampling temperature being 0.6, one word after another (i.e. the model was not influenced by the earlier-generated words). Some of the words were not successfully translated for the first time, since they were too common like \emph{what} and the model did not understand they are actually words that need to be translated. The failed ones were gathered in separated files that can be retrieved in Supplementary information and accompanied by more explicit prompting (\emph{Wie kann man ``was" übersetzen?} 'How can we translate the word ``what"?'). Czech pseudowords were generated on 2023/12/7, German pseudowords were generated on 2024/2/12 by the same version of the model (gpt-4-0613). The resulting words were then transcribed into Czech and German orthography so that they could be presented to the experiment participants. The transcription rules are provided in full in the protocol. Since there is practically no reasonably short string of Latin alphabet characters that is never used on the Internet, we randomly checked the newly created words for meaning. The given strings had only marginal number of occurrences and they did not refer to the given meaning.

\subsubsection{Creating the word lists}
The two lists of the 68 pseudowords --- German-prompted pseu\-do\-words and Czech-prompted pseudo\-words --- were used to create four versions of the experiment, since we wanted to present both pseudoword lists to both Czech and German speakers. Therefore, we had:

\begin{itemize}
    \item Czech-prompted pseudowords with Czech trans\-lations for Czech speakers
    \item Czech-prompted pseudowords with German trans\-la\-tions for German speakers
    \item German-prompted pseudowords with German trans\-la\-tions for German speakers
    \item German-prompted pseudowords with Czech trans\-la\-tions for Czech speakers
\end{itemize}

The lists were presented in an online experiment, either on android tablets and distributed in person among participants, or distributed by a link.

\subsection{Participants}
\subsubsection{Human participants}
Participants were recruited either in person, distributing android tabled with BlackSquare appli\-ca\-tion version of the experiment, or by distributing the link to the experiment through online platforms, such as Facebook and university webpages. In total the data of 672 participants who finished the experiment was collected across all conditions. Demographical details of the participants can be seen in Table \ref{tab:participants}.

\begin{table*}[t]
\centering
\caption{Summary of Participant Statistics.}\label{tab:participant_statistics}
\label{tab:participants}
\tabcolsep=0pt

\begin{tabular*}{0.7\textwidth}{@{\extracolsep{\fill}}llccc@{\extracolsep{\fill}}}
\toprule%
Participants & Dataset & Count & Female ratio (\%) & Average Age \\
\midrule
Czechs  & \multirow{2}{*}{Czech pseudowords} & 220 & 66.82 & 30.25 \\
Germans &                                       & 154 & 67.53 & 25.15 \\
Czechs  & \multirow{2}{*}{German pseudowords} & 176 & 67.05 & 26.16 \\
Germans &                                       & 122 & 32.79 & 29.74 \\
\midrule
\textbf{Total} &                                & 672 & 60.86 & 27.92 \\
\botrule
\end{tabular*}
\end{table*}

\subsubsection{LLM participants}
To fully examine the pseudowords GPT-4 created, we wanted to see if the LLMs themselves will have a better chance to correctly assign the pseudowords to the meanings. We simulated both Czech participants guessing Czech pseudowords in Czech transcription and German participants guessing German words in German transcription.

We tested all combinations of words (i.e., $68(68-1)/2=2278$ combinations for each language). To simulate participants, we selected gpt-4-0613, the same model used to generate the words. For comparison, we also tried another state-of-the-art model from a competing company, Anthropic's Claude 3.5 Sonnet (version claude-3-5-sonnet-20241022). Since the prompting was zero-shot, we ended up with:

\begin{itemize}
    \item 2278 GPT-4 Czech participants guessing Czech-prompted pseudowords
    \item 2278 claude-3-5-sonnet Czech participants gues\-sing Czech-prompted pseudowords
    \item 2278 GPT-4 German participants gues\-sing German-prompted pseudowords
    \item 2278 claude-3-5-sonnet German partici\-pants gues\-sing German-prompted pseudowords
\end{itemize}

\subsection{Procedure}
\subsubsection{Human participants}
Before the beginning of the experiment, the following instruction for Czech speakers was displayed (in Czech) on the screen: `You are currently participating in a linguistic experiment. Your responses will be processed for scientific purposes. On each screen, you will see four words: two Czech words and their translations from an unknown language. Your task is to correctly match the words (using finger gestures or mouse draws), even though you do not know the foreign language. There are correct answers. Use your linguistic intuition. You may withdraw from the experiment at any time, in which case your responses will be deleted.' The same instruction for German speakers was displayed in German (with ``two German words" instead of ``two Czech words"). Participants were required to tap the OK button to proceed.

\begin{figure}[ht]
\centering
\begin{tikzpicture}[scale=0.88]
 \node[minimum width=1cm, minimum height=1cm] at (0,0) (ul1) {den};
 \node[minimum width=1cm, minimum height=1cm] at (0,-2) (dl1) {gurúm};
 \node[minimum width=1cm, minimum height=1cm] at (2,0) (ur1) {hora};
 \node[minimum width=1cm, minimum height=1cm] at (2,-2) (dr1) {líja};
 \draw (ul1.north west) rectangle (dr1.south east);
 
 \begin{scope}[xshift=5cm]
   \node[minimum width=1cm, minimum height=1cm] at (0,0) (ul2) {den};
   \node[minimum width=1cm, minimum height=1cm] at (0,-2) (dl2) {gurúm}; 
   \node[minimum width=1cm, minimum height=1cm] at (2,0) (ur2) {hora};
   \node[minimum width=1cm, minimum height=1cm] at (2,-2) (dr2) {líja};
   \draw (ul2.north west) rectangle (dr2.south east);
   \draw[-,ultra thick] (ul2) -- (dr2);
   \draw[-,ultra thick] (ur2) -- (dl2);
 \end{scope}
 
 \draw[->,ultra thick] (3,-1) -- (4,-1);
\end{tikzpicture}
\caption{An example of an experiment trial. The upper pair of words is in Czech (the first word means \emph{day}, the second one means \emph{mountain}), and the lower pair are matching Czech-prompted pseudowords.}
\label{fig:procedure}
\end{figure}
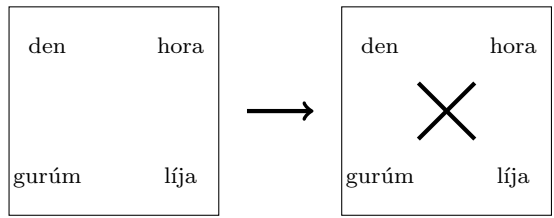

Each participant received 34 assignments, each consisting of a set of four words: the upper pair in Czech/German and the lower pair were the two matching pseudowords. The order of the pairs as well as the pairing of the words was randomized for each participant. Participants were tasked with matching the corresponding words using gestures (see Figure \ref{fig:procedure}). After completing the experiment, participants were informed about their results and asked to provide information about their age and gender.

\subsubsection{LLM-based participants}
A modified version of the experiment was presented to the LLMs (Claude 3.5 Sonnet and GPT-4). All categories were prompted in a way that closely resembled the circumstances of the human experiment. the only exception was that the prompt included a call to reflect before answering, as we again wanted to know the motivations that led the models to assign meanings. 

In both models, we set the sampler temperature at 0.0 (deterministic outputs). All experiments were conducted on 2024/11/5.

An example of a specific prompt and response from Claude 3.5 Sonnet is in the Table \ref{tab:german_english}.

\begin{table*}[t]
\caption{German Prompt for Claude 3.5 Sonnet and the Response.}\label{tab:german_english}
\tabcolsep=0pt
\begin{tabular*}{\textwidth}{@{\extracolsep{\fill}}p{0.45\textwidth}p{0.45\textwidth}@{\extracolsep{\fill}}}
\toprule%
\textbf{German} & \textbf{English translation} \\
\midrule
\textbf{System prompt:} & \textbf{System prompt:} \\
Sie nehmen gerade an einem linguistischen Experiment teil; Ihre Antworten werden für wissenschaftliche Zwecke verarbeitet. 
Auf jede Folie werden Ihnen 4 Worte angezeigt. 2 deutsche Worte sind Übersetzungen von 2 Worten aus einer unbekannten Sprache. 
Ihre Aufgabe ist es, die Worte korrekt zuzuordnen, auch wenn Sie die Fremdsprache nicht kennen. Es gibt richtige Antworten. 
Nutzen Sie Ihre sprachliche Intuition. Denken Sie zuerst darüber nach, wie jedes Wort klingt. \newline
Ordnen Sie dann zu, welches Wort zu welchem gehört, im folgenden Format: \newline
entweder 1-2;2-1 \newline
oder 1-1;2-2 &
You are participating in a linguistic experiment; your answers will be processed for scientific purposes. 
Each slide will display 4 words. 2 German words are translations of 2 words from an unknown language. 
Your task is to correctly match the words, even if you don’t know the foreign language. There are correct answers. 
Use your linguistic intuition. First, think about how each word sounds. \newline
Then match the words in the following format: \newline
either 1-2;2-1 \newline
or 1-1;2-2 \\
\midrule
\textbf{User prompt:} & \textbf{User prompt:} \\
Deutsches Wort 1: ungreifbar \newline
Deutsches Wort 2: Bewegung \newline
Fremdes Wort 1: luhoomi \newline
Fremdes Wort 2: looana &
German Word 1: intangible \newline
German Word 2: movement \newline
Foreign Word 1: luhoomi \newline
Foreign Word 2: looana \\
\midrule
\textbf{Response:} & \textbf{Response:} \\
Lassen Sie mich die Wörter nach ihrem Klang analysieren: \newline
``ungreifbar" - ein abstraktes, flüchtiges Konzept \newline
``luhoomi" - klingt weich, schwebend, schwer zu fassen \newline
``Bewegung" - ein dynamisches, konkretes Wort \newline
``looana" - klingt fließend, dynamisch \newline
Basierend auf den Klangqualitäten würde ich folgende Zuordnung vorschlagen: \newline
1-1;2-2 \newline
Also: \newline
ungreifbar - luhoomi \newline
Bewegung - looana \newline
Die weichen, schwebenden Laute in ``luhoomi" passen gut zum Begriff ``ungreifbar", während die fließenden Laute in ``looana" gut zur ``Bewegung" passen. &
Let me analyze the words based on their sound: \newline
``intangible" - an abstract, fleeting concept \newline
``luhoomi" - sounds soft, floating, hard to grasp \newline
``movement" - a dynamic, concrete word \newline
``looana" - sounds flowing, dynamic \newline
Based on the sound qualities, I suggest the following match: \newline
1-1;2-2 \newline
That is: \newline
intangible - luhoomi \newline
movement - looana \newline
The soft, floating sounds in ``luhoomi" align well with the concept of ``intangible," while the flowing sounds in ``looana" match ``movement" well. \\
\botrule
\end{tabular*}
\end{table*}

\section{Results and discussion}
Our project generated an extensive dataset that enables testing numerous hypotheses. However, within the scope constraints of this paper, we focus exclusively on our initial research questions. The complete raw dataset, including all scripts used for pseudoword generation, experiment execution, and analysis, is available at \url{https://osf.io/ywjrk/}. This repository also includes a comprehensive 115-page \emph{Laboratory Protocol} documenting the experimental procedures and analyses in detail, along with additional statistical analyses beyond those presented here. We encourage researchers interested in exploring other hypotheses to utilize this publicly available dataset and accompanying materials.

\subsection{Correctness overview}
\begin{figure}[!htbp]
    \centering
    \includegraphics[width=0.48\textwidth]{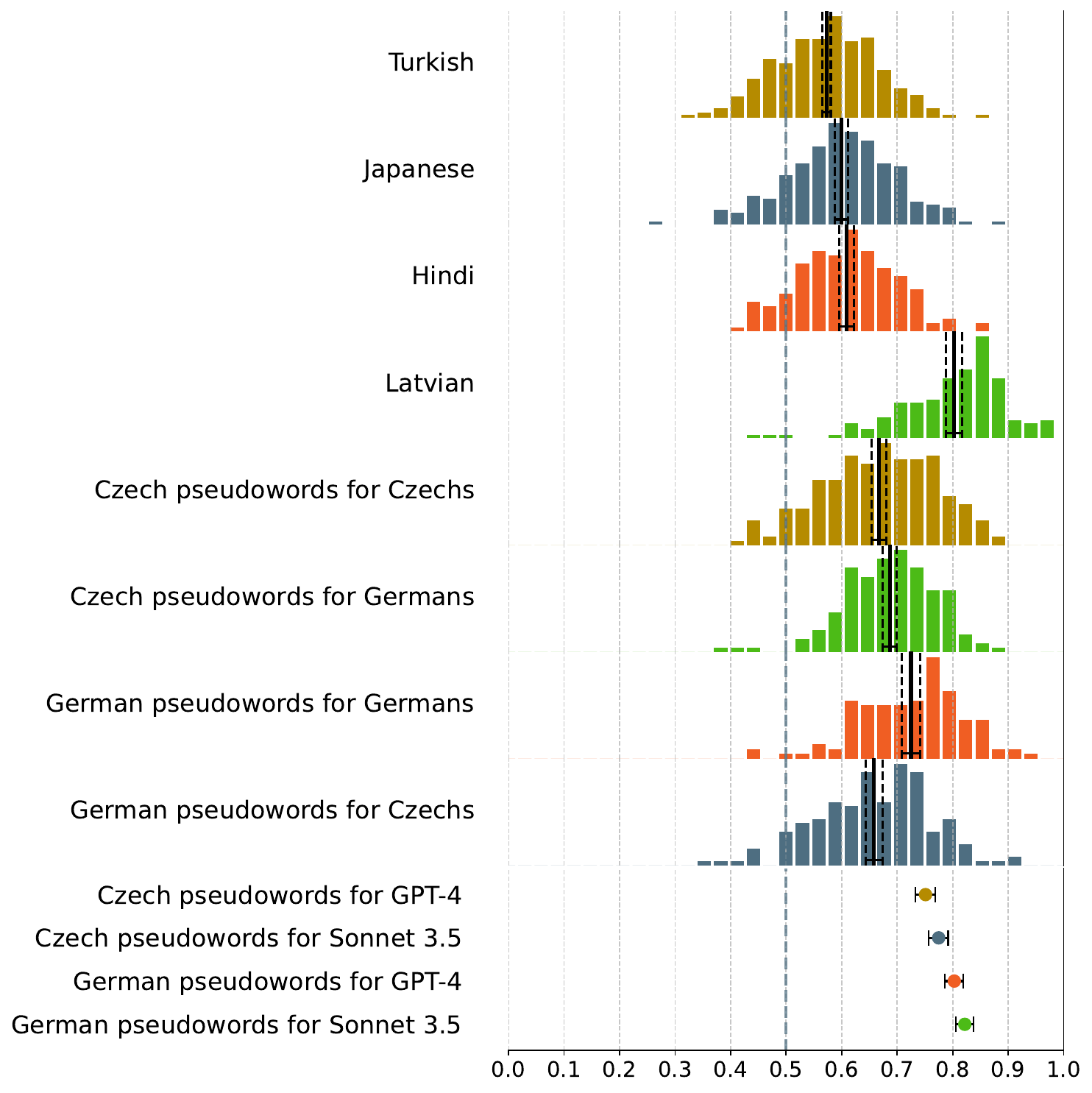}    
    \caption{The distribution of the correctness of the answers.}
    \label{fig:correctness_overall}
\end{figure}

\begin{figure}[!htbp]
    \centering
    \includegraphics[width=0.48\textwidth]{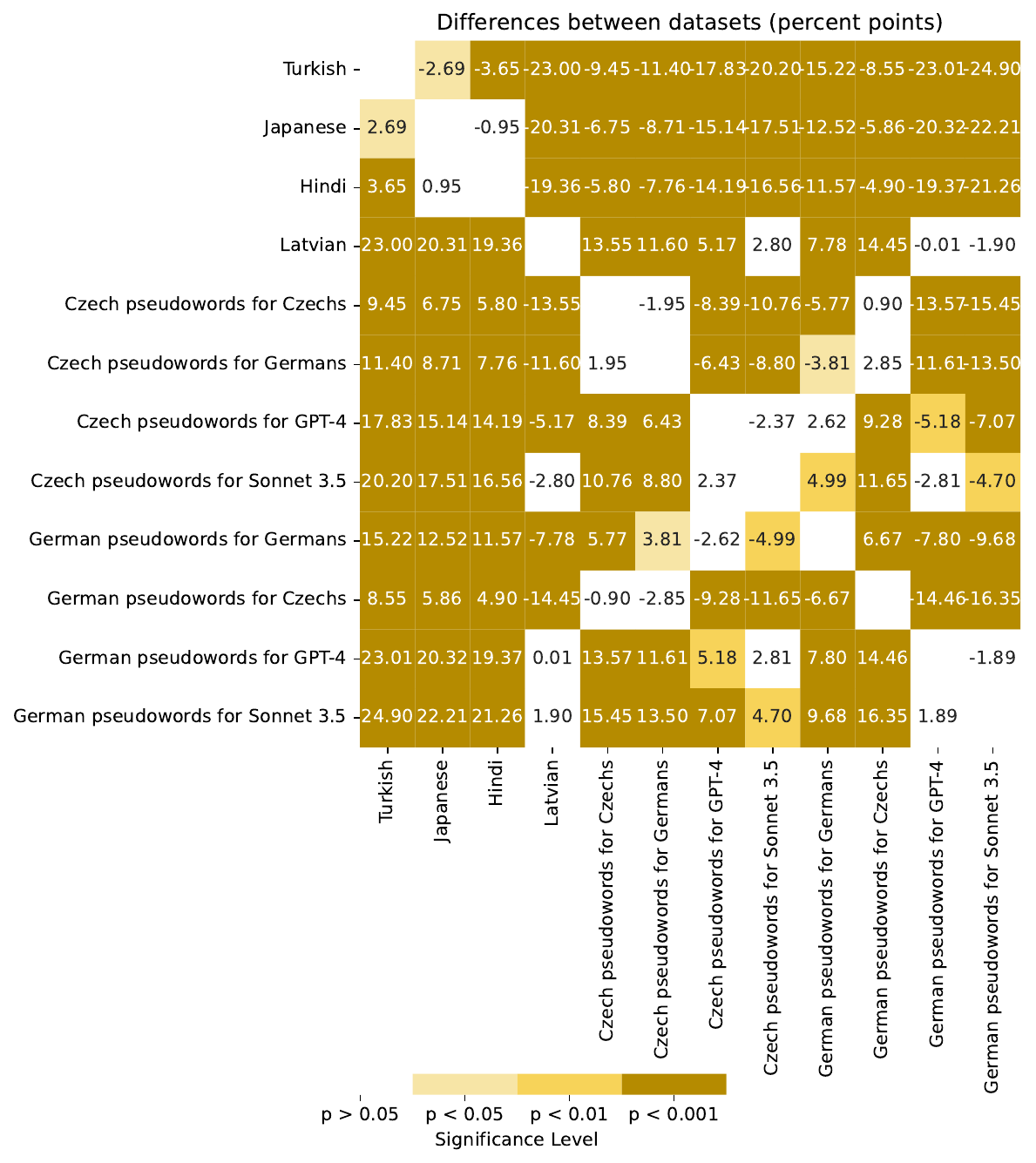}    
    \caption{The comparison of the correctness of the answers across all datasets.}
    \label{fig:copmarison_correctness}
\end{figure}

First, let us examine the basic correctness statistics. We present these results alongside findings from \citet{NasClanekVeSpeechi} to compare correctness rates between natural languages and generated pseudowords. The proportion of correct responses for each observed group is best illustrated in Table \ref{tab:dataset_comparison_correctness}. The 95\% confidence intervals were calculated using bootstrap resampling grouped by participants (as each participant responded to an ensemble of questions). For LLM-based participants, we employed binomial confidence intervals, as individual trials were mutually independent. For visual clarity, these values and their histograms are also presented in Figure \ref{fig:correctness_overall}. The differences between datasets (measured in percentage points) and statistical significance of the differences are presented in Figure \ref{fig:copmarison_correctness}.

Both the table and the charts demonstrate that human participants guessing the artificially generated language consistently perform significantly better than those guessing unfamiliar non-generated languages. The sole, albeit predictable, exception is Latvian, which, while not mutually intelligible with Czech, is related to it and contains numerous cognates, resulting in correctness exceeding 82\%. Excluding Latvian, the highest performance for natural languages was observed in Hindi, where Czech speakers achieved accuracy of 60.9 \% (upper confidence interval limit 62.3\%). The lowest artificial language performance was for the German-prompted pseudowords, which Czech speakers guessed with a correctness of 65.8\% (lower CI limit 64.6\%). 
It suggests that LLMs successfully used iconicity when creating the pseudowords, and this iconicity cues helped the speakers to connect the meanings with corresponding forms.

The data also reveals that our sample of participants performed slightly better with artificial languages prompted in their native language --- Czech speakers better guessed the Czech-prompted language, and German speakers the German-prompted language --- though this difference is statistically significant only for German speakers. German speakers performed significantly better than Czech speakers on German dataset (difference being 6.67 percent points, 95 \% CI 4.41--8.91). Our sample of Germans was also better than our sample of Czechs in guessing Czech-prompted pseudowords, but the difference is not significant under Bonferoni correction (a difference of 1.95 percentage points, with a non-corrected 95\% CI of 0.063--3.835). These inconclusive results suggest that LLMs rely not only on language-specific iconicity but also on certain general features --- features that are general enough to make sense cross-linguistically, at least within the Czech-German context.

The most striking difference, however, is between human participants and LLM-based participants, with a gap of approximately ten percentage points that is statistically significant across all cases. The differences between guessing models is not statistically significant (in our sample, somewhat surprisingly, Claude 3.5 Sonnet achieves better results than GPT-4, which was the source of the generated pseudowords, but these differences are small and not statistically significant). 

What is statistically significant, however, is that both models perform better at guessing words generated in German compared to those generated in Czech --- for GPT-4, the difference is 5.180 percent points (CI: 2.766--7.594) For Claude 3.5 Sonnet, the difference is 4.697 percent points (CI: 7.024--2.371), which is probably because the training data contain more German texts. Alternatively, The overall better scores across participants groups for German-prompted pseudowords can be caused by untrackable idiosyncracies of the particular two pseudowords datasets used in this experiment, not with German and Czech langauge in particular. For that, replication of the study with newly generated Czech-prompted and German-prompted pseudoword could be interesting.

\begin{table}[htbp!]
\caption{Comparison of correctness of the answers --- the human and LLM-based participants.}
\centering
\resizebox{\columnwidth}{!}{
\begin{tabular}{llcc}
\toprule
\textbf{Dataset} &\textbf{Participants} & \textbf{Average} & \textbf{CI (95\%)} \\
\midrule

Turkish words & \multirow{4}{*}{\makecell{Czech}} & 0.573 & 0.565--0.580 \\
Japanese words                                       &  & 0.600 & 0.588--0.612 \\
Hindi words                                       &  & 0.609 & 0.596--0.623 \\
Latvian words                                      &  & 0.803 & 0.788--0.817 \\
\midrule

\multirow{4}{*}{\makecell{Czech prompted\\pseudowords}} & Czechs & 0.667 & 0.657--0.678 \\
                                       & Germans & 0.687 & 0.674--0.699 \\
                                       & GPT-4 & 0.751 & 0.733--0.769 \\
                                       & Sonnet 3.5 & 0.775 & 0.757--0.792 \\
\midrule
\multirow{4}{*}{\makecell{German prompted\\pseudowords}} & Germans & 0.725 & 0.711--0.739 \\
                                        & Czechs & 0.658 & 0.646--0.670 \\
                                        & GPT-4 & 0.803 & 0.786--0.819 \\
                                        & Sonnet 3.5 & 0.822 & 0.805--0.837 \\
\botrule
\end{tabular}
}
\label{tab:dataset_comparison_correctness}
\end{table}

\subsection{Logistic regression}
We are interested not only in participants' success rates but also in understanding the factors that influenced their performance. What independent variables played a role in determining the meanings of unknown words, what features participants relayed on? We can draw on results from previous comparable research conducted with natural languages \citep{NasClanekVeSpeechi}, where we found that people relied primarily on \emph{vector phonological similarity}, \emph{edit phonological similarity}, and \emph{length agreement}. In the previous study, we also included variables named \emph{part of speech mismatch}, \emph{semantic distance}, and \emph{trial order}. For human participants, we also incorporated demographic variables of \emph{age} and \emph{gender}. These demographic variables were not applicable for LLM-based participants, as well as \emph{trial order}, since stimuli were presented independently rather than in sets.

For detailed operationalization of these variables, we refer readers to the \emph{Protocol} (\url{https://osf.io/ywjrk/}), which contains complete results, variable distributions, and graphs showing the relationship between Correctness and these variables. Here, we provide a brief description:
 
\emph{Vector phonological similarity} measures the similarity between natural language words and generated words, with the similarity metric based on the frequency of phonemes important for iconicity according to \citet{ThompsonVanHoeyDo2021}. \emph{Edit phonological similarity} is analogous but uses edit distance weighted for phoneme similarity, making it more general and less specifically related to iconicity. \emph{Length agreement} is a metric indicating whether, in a given quaternity of words, short words in the natural language are translated by short generated words and long by long ones (positive value), or vice versa (negative value).

\emph{Part of speech mismatch} determines whether two given words in a quartet belong to the same part of speech. \emph{Semantic distance} measures how far apart the two natural language words in a quaternity are in the embedding space (based on embeddings taken from \citet{Fares2017word} for both Czech and German). \emph{Trial order} simply indicates which task number an assignment was for a given participant, assuming they either improved with practice or became more fatigued over time.

\emph{Inverse trial} is a somewhat technical variable examining whether participants tended to match words in parallel (as presented) or crosswise (changing the order).

For human participants, we employed a mixed logistic regression model with \emph{correctness} as the target variable. The random components included \emph{part of speech mismatch}, \emph{semantic distance}, \emph{edit phonological similarity}, \emph{vector phonological similarity}, \emph{length agreement}, \emph{inverse trial}, and \emph{trial order}, effectively encompassing all variables except \emph{gender} and \emph{age}. All co-variables were standardized. For LLM-based participants, since participant-related demographic variables were not applicable, we used simple logistic regression.

In both cases, we utilized the \emph{GAMLj} package \citep{Gallucci2024a} within the \emph{Jamovi} statistical software \citep{jamovi2024}. All detailed settings are documented in the Protocol and are available in the Jamovi files included in the Supporting Data.

The model results are visualized in the figures (\ref{ModelEstimate} for human participants) and (\ref{ModelLLMEstimate} for LLM-based participants). We will now examine each variable in detail.

\newcommand{\ChartModel}[1]{
\begin{figure}[p]
\centering
    \includegraphics[width=0.42\textwidth]{Supplementary_information/regression/gpt4ces-ces_Correctness_of_the_answer_#1_bar_plot}
    \includegraphics[width=0.42\textwidth]{Supplementary_information/regression/gpt4ger-ger_Correctness_of_the_answer_#1_bar_plot}
    \includegraphics[width=0.42\textwidth]{Supplementary_information/regression/gpt4ces-ger_Correctness_of_the_answer_#1_bar_plot}
    \includegraphics[width=0.42\textwidth]{Supplementary_information/regression/gpt4ger-ces_Correctness_of_the_answer_#1_bar_plot}    
\caption{Model of correctness rate (human participants).}
\label{Model#1}
\end{figure}
}

\newcommand{\ChartModelLLM}[1]{
\begin{figure}[!pbt]
\centering
    \includegraphics[width=0.49\textwidth]{Supplementary_information/regression/gpt4ces-gpt4ces_Correctness_of_the_answer_#1_bar_plot}
    \includegraphics[width=0.49\textwidth]{Supplementary_information/regression/gpt4ger-gpt4ger_Correctness_of_the_answer_#1_bar_plot} 
    \includegraphics[width=0.49\textwidth]{Supplementary_information/regression/gpt4ces-sonnet35ces_Correctness_of_the_answer_#1_bar_plot}
    \includegraphics[width=0.49\textwidth]{Supplementary_information/regression/gpt4ger-sonnet35ger_Correctness_of_the_answer_#1_bar_plot}   
\caption{Model of correctness rate (large language model based participants).}
\label{ModelLLM#1}
\end{figure}
}

\ChartModel{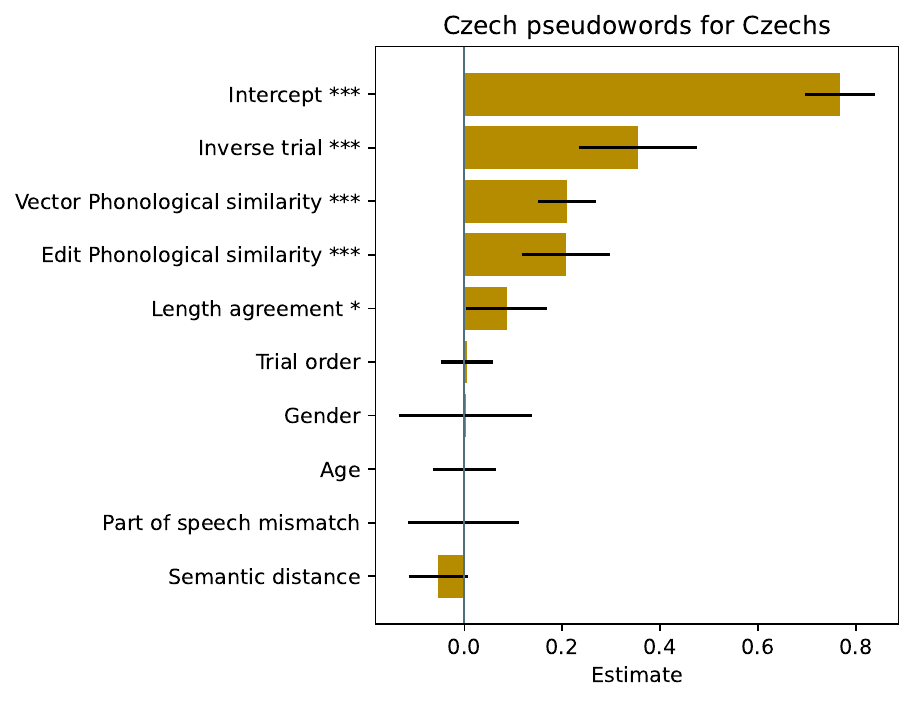}
\ChartModelLLM{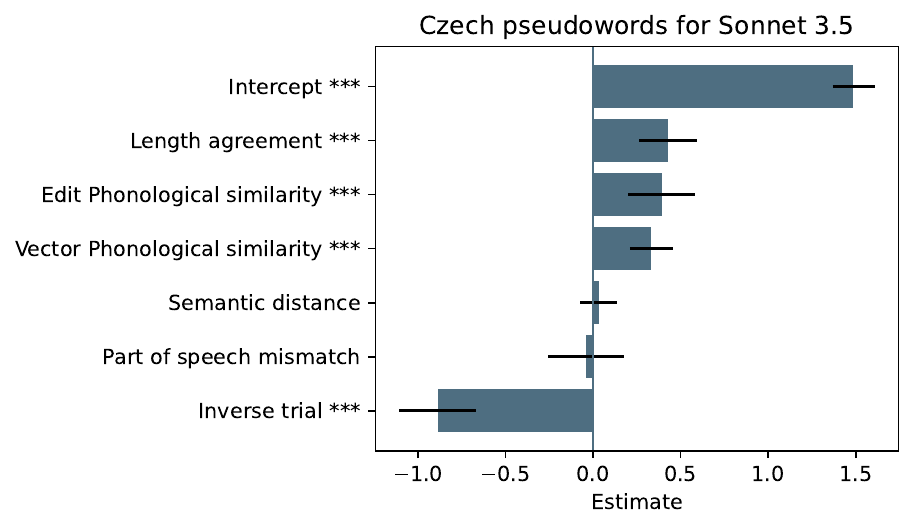}

Let us first examine the last mentioned dependent variable, \emph{inverse trial}. Although trials were presented to participants in random order, bias toward either direction did not help them --- on the contrary, it would have made more sense to distribute their guesses evenly. However, neither group did this: human participants tended to make crosswise associations, while LLM-based participants tended to match the first mentioned word with the first mentioned pseudoword. As noted, this variable is primarily technical in nature and has no relationship to iconicity.

In contrast, the most iconicity-related variable under study is \emph{vector phonological similarity}, which plays a significant role across all datasets except for German-prompted pseudowords presented to Germans and GPT-4 based participants. \emph{Edit phonological similarity} plays a similarly important role (except for Czech-prompted pseudowords presented to Germans), a variable that appears to capture similar effects, although these variables do not strongly correlate with each other (the lowest correlation is found in Czech pseudowords presented to Czech participants: $\rho=-0.69$, the highest in German pseudowords presented to Czech participants $\rho=0.655$). This variable was significant across all datasets except Czech-prompted pseudowords presented to Germans.

A similar effect was shown for \emph{length agreement}. This variable proved significant across all datasets and was even the most important factor after the intercept in several cases (specifically, in all datasets where participants were LLMs). 

The hypothesis that \emph{semantic distance} would have an influence seems quite plausible --- it is typically more challenging to distinguish between words that are semantically close --- however, this was only confirmed for German prompted pseudowords for Czechs and Czech prompted pseudowords for GPT-4 (thus showing no systematic pattern).

Similarly unsystematic results can be observed for \emph{parts of speech mismatch}: we would expect a positive influence, meaning that word pairs differing in part of speech would be easier to guess. However, this was only confirmed for Czech stimuli for GPT-4, while the opposite effect was found for German stimuli both for Czechs and Germans, and it proved non-significant in all other cases.

Demographic variables (gender and age) played no significant role in any dataset, and the same was true for trial order.

To summarize, the three most influential features influencing correct guesses of words from natural languages (length agreement, edit phonological distance, and vector phonological distance) was found equally important when guessing pseudowords, and they were important not only for human participants but also for LLM-based participants. We interpret it as an evidence that LLMs are not just able to generate iconic pseudowords, but that they are able to replicate clues helping connect forms and meanings in natural languages.

Overall, the large intercepts indicate that there are still many variables we have not managed to capture in our analysis.

\subsection{LLMs' justification of decisions}
As mentioned in the previous chapter, logistic regression revealed a large intercepts, that means, there might be some hidden factors influencing the correctness of the guesses. These factors are not related to the phonetic resemblances or length agreements. To uncover some of the possible factors, we explored the answers of LLM-based participants. As described in the Methodology chapter, these participants were asked to explain their decision when pairing the words with their translations, and while we do not have space in this paper to quantify the explanations they provided, we can describe them qualitatively. We will use the translated examples from Czech prompts and outputs, however, the models behaved comparably in German.

Firstly, we will describe the reasoning\footnote{\emph{Reasoning} is not used as a strictly defined term here, just as the \emph{thoughs} in the notion of a \emph{chain of thoughts} are, strictly speaking, quite different from \emph{thoughs} as understood in cognitive science.} of GPT-4 while guessing the pseudowords. Unsurprisingly, it frequently reasoned based on word length, e.g., ``\textit{opět} `again' is short and simple, so it could correspond to the foreign word \textit{éró}, which is also short and simple. On the other hand, \textit{pokrývka} `cover' is a longer word with more syllables, which could correspond to the foreign word \textit{mufutu}, which is also longer and has more syllables." Simplicity/complexity of the words was another feature often mentioned (e.g., ``\textit{šusmin} sounds like a more complex word, which could correspond to a more complex concept, such as \textit{prodavač} `shop assistant'").

The model frequently reasoned 'intuitively' and described feelings about the sound of the words connected to particular meanings, e.g., ``The word \textit{otěže} `reins' feels like something that could be translated as \textit{grukalé}. It has that old, rustic sound that might be related to equestrian equipment. On the other hand, \textit{prodavač} `shop assistant' sounds like something more modern and less rustic, which might correspond to the word \textit{šusmin}."

The model usually introduced such explanations by a phrase \textit{připadá mi} to `it seems to me'. From these intuitive feelings about words, the characteristics of something sounding tough/hard and something sounding soft/gentle were mentioned very often (e.g., ``The first thing that comes to my mind is that \textit{grúom} sounds harder and \textit{nalüta} softer. \textit{Práce} `work' is something that can be hard, while \textit{směr} `direction' is something softer, more abstract"). The abstract/concrete nature of the words was mentioned frequently, as well as negative/positive aura. The model mentioned, e.g., that a word has commercial sound, is warm, sounds formal, exotic, aggressive, light, or playful.

In the case of GPT-4, we explored if the argumentation the model gave when answering agreed with the explanations during the words' creation. Since the creation and the answering was done on the same model, we expected an overlap in reasoning. However, the process how the model created the words seems to differ from the process of guessing the meanings. The argumentation in words creation was more compositional, the morphemes were often explained separately, such as ``The word for grandmother in the language of the Tubar tribe could be \textit{Lumiru}. \textit{Lu} could represent softness and tenderness, while \textit{miru} could symbolize wisdom and experience." While the model used abstract characteristics of the sounds and meanings such as softness/hardness, friendliness, wisdom, etc., in argumentation for both word creation and guessing, it did not use abstract/concrete nature or negative/positive aura of the words during the word creations. Length and complexity were also not used as explicit explanations for the word creation (i.e., it did not explain that it created a long Tubar word for a comparably long Czech word). While answering, the model contemplated about the word classes (e.g., ``\textit{Tukalu} feels more like a noun because it has more syllables and is longer, which is common for nouns in Czech"), which was not a characteristic mentioned during word creation.

From our observations, the model did not use the same reasoning when guessing the word meanings as when creating them. As an example, consider the word \textit{východní} `eastern'. During the creation, the model described the reasoning as follows:``In the language of the Tubar tribe, \textit{východní} `eastern' could evoke sounds associated with the sunrise, such as the chirping of birds, the rustling of leaves in the breeze, or the quiet sound of dawn. Therefore, the word could contain sounds like \textit{tss}, \textit{shh}, or \textit{ee}, which could resemble these sounds. The word for eastern in the language of the Tubar tribe could thus be \textit{Sheteshee}." In contrast, the explanations GPT-4 offered when guessing were , e.g., that the word is long and complicated (or short and simple); it has more syllables and is more complex than the other word; it has an `eastern sound', as if it was from a Semitic language; it feels exotic as if from eastern culture; it is gentle and not aggressive; it sounds like something to do with direction; etc..

From this qualitative inspection, we extracted the information that 1) next to length and phonological similarity, the GPT-4 model used complexity of the words and the `intuition' about how they sound (e.g., rough, aggressive, playful) for the guessing, and 2) the model used fairly different reasoning for word creation and for word guessing.

An interesting finding occurred when we inspected the reasoning of Sonnet-3.5. In this case, we could not compare the word creation and guessing processes, since the words were created by GPT-4. Therefore, we explored the reasoning for word guessing only.

While Sonnet-3.5 used mostly similar strategies as GPT-4 during guessing, we discovered that it differed from the reasoning offered by GPT-4 in several aspects. Most notably, Sonnet-3.5 argued more often on the level of phonemes and morphemes (e.g., ``\textit{Babička} `grandmother' --- a longer word, with a softer sound due to the \textit{b} and \textit{č}. \textit{Pulunu} --- contains more \textit{u}, sounds heavier"). Additionally, it used them to assess parts of speech (e.g., ``\textit{grukalé} --- this word sounds more complex, it has more syllables and ends in \textit{-é}, which could evoke an adjective or adverb. \textit{Gorúú} --- this word is shorter and has a distinct ending \textit{-úú}, which might resemble a noun"). Paradoxically, logistic regression has shown that \emph{parts of speech mismatch} was not utilized by Claude 3.5 Sonnet significantly (see Figure \ref{ModelLLMEstimate}). When we look at the argumentation for the guessing of the word \textit{východní} `eastern', we find such arguments as ``it sounds sharper and contains \textit{š}", or ``it contains sibilants and soft phonemes", etc. In this regard, Sonnet-3.5 was closer to the reasoning of GPT-4 during word creation than when we compared GPT-4 during guessing with GPT-4 during word creation. This fact might have caused the slightly better correctness rate in Sonnet-3.5 than in GPT-4 in respective datasets. However, since the models did not differ significantly, and since this chapter focused on qualitative observations, it can serve only as an insight into possible factors influencing the correctness of guesses. 

It is worth noting that the models often made factual mistakes in the reasoning, such as that they argued that some word has two syllables when it had three syllables, or identified consonant /š/ as a sharp consonant.

\newcommand{\ChartComparison}[4]{
\begin{figure*}[!tbp]
\centering
    \includegraphics[width=0.5\textwidth]{Supplementary_information/individual_words/Scatter-#1-GeneratedX.pdf}\includegraphics[width=0.5\textwidth]{Supplementary_information/individual_words/Scatter-#1-Meaning.pdf}
    \\
    \includegraphics[width=0.5\textwidth]{Supplementary_information/individual_words/Scatter-#2-GeneratedX.pdf}\includegraphics[width=0.5\textwidth]{Supplementary_information/individual_words/Scatter-#2-Meaning.pdf}    
\caption{#4 Left column shows the pseudowords, Right column shows their English meanings.}
\label{ChartComparison_Words_#3}
\end{figure*}
}

\subsection{Comparison at the word level}

\ChartComparison{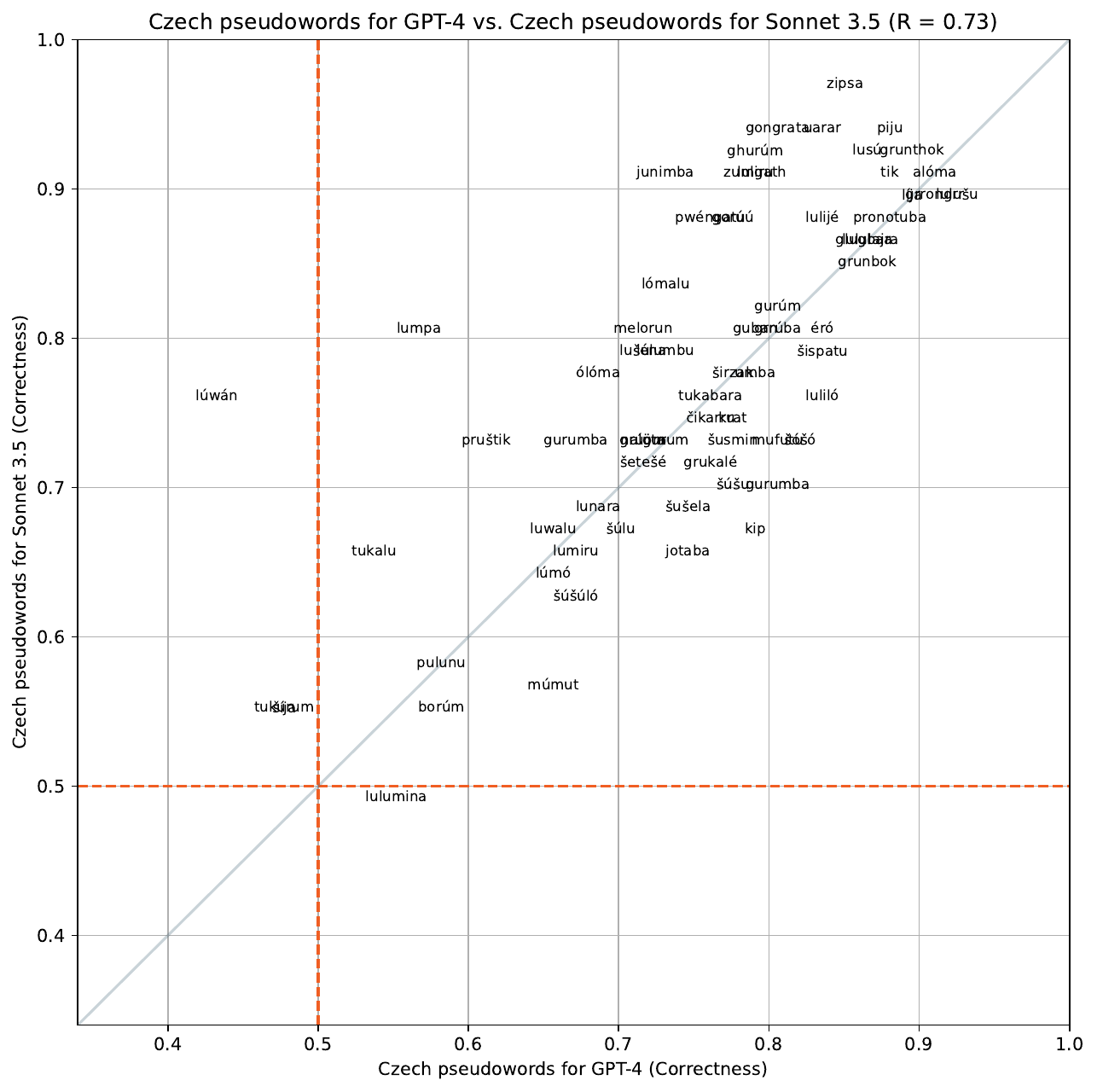}{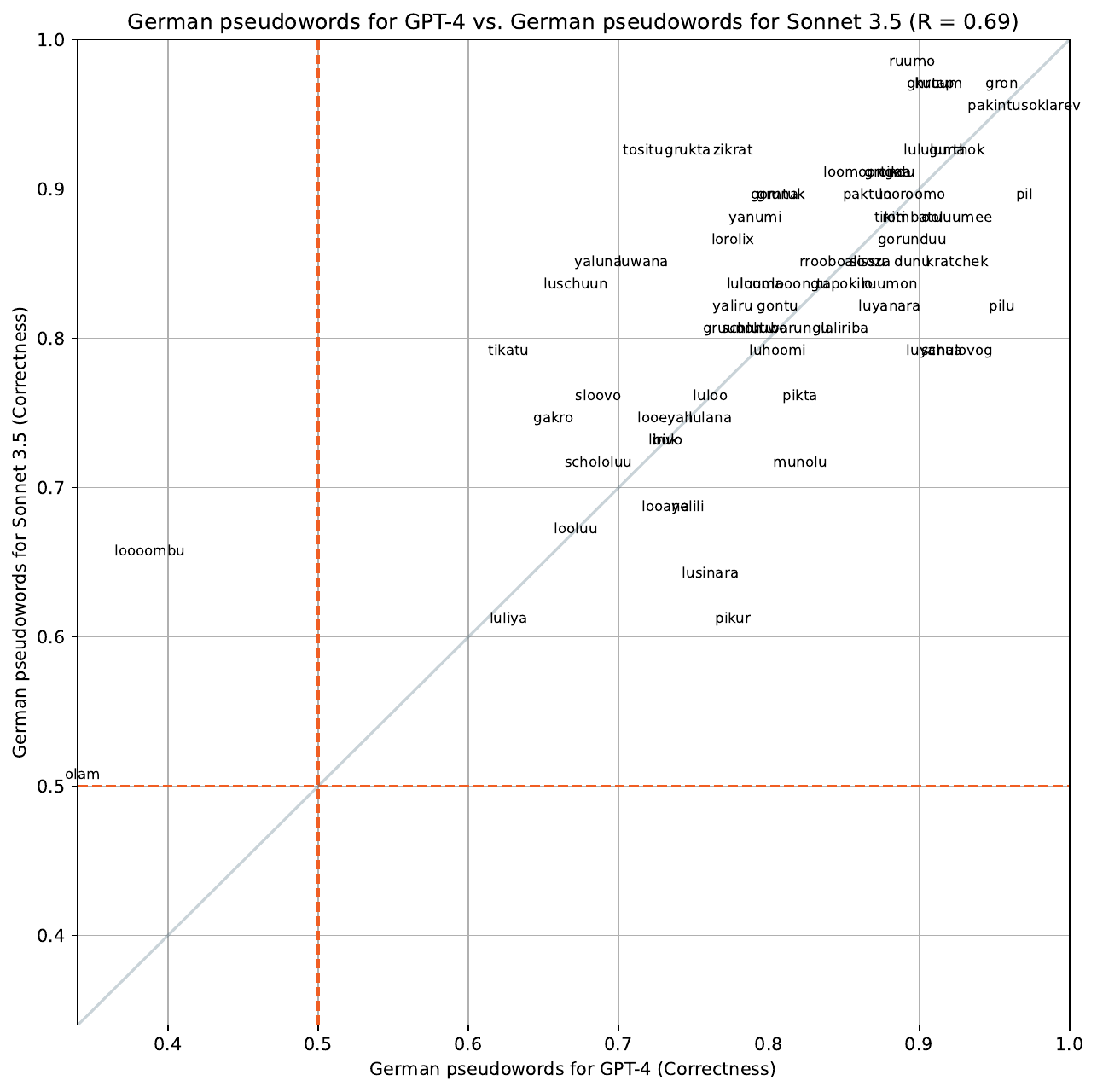}{llm-llm}{Czech prompted pseudowords presented to GPT-4 and Claude Sonnet 3.5 based participants (top), and German prompted pseudowords presented to GPT-4 and Claude Sonnet 3.5 based participants (bottom).}

Given that the models differ in how they explain their word-matching strategies, we might expect differences in which words they correctly or incorrectly assign. As shown in Figure \ref{ChartComparison_Words_llm-llm}, there are indeed words that GPT-4-based participants guessed correctly while Claude Sonnet 4.5 based participants guessed incorrectly (e.g., \textit{pikur} `which' in the German dataset, Figure \ref{ChartComparison_Words_llm-llm}, bottom charts) and vice versa (e.g., \textit{loooombu} `a lot' in the German dataset). However, we also find words where they agreed in their judgments: for instance, both models systematically guessed \textit{gron} `he was' correctly, while both consistently failed to correctly identify \textit{luliya} `contact'.

\ChartComparison{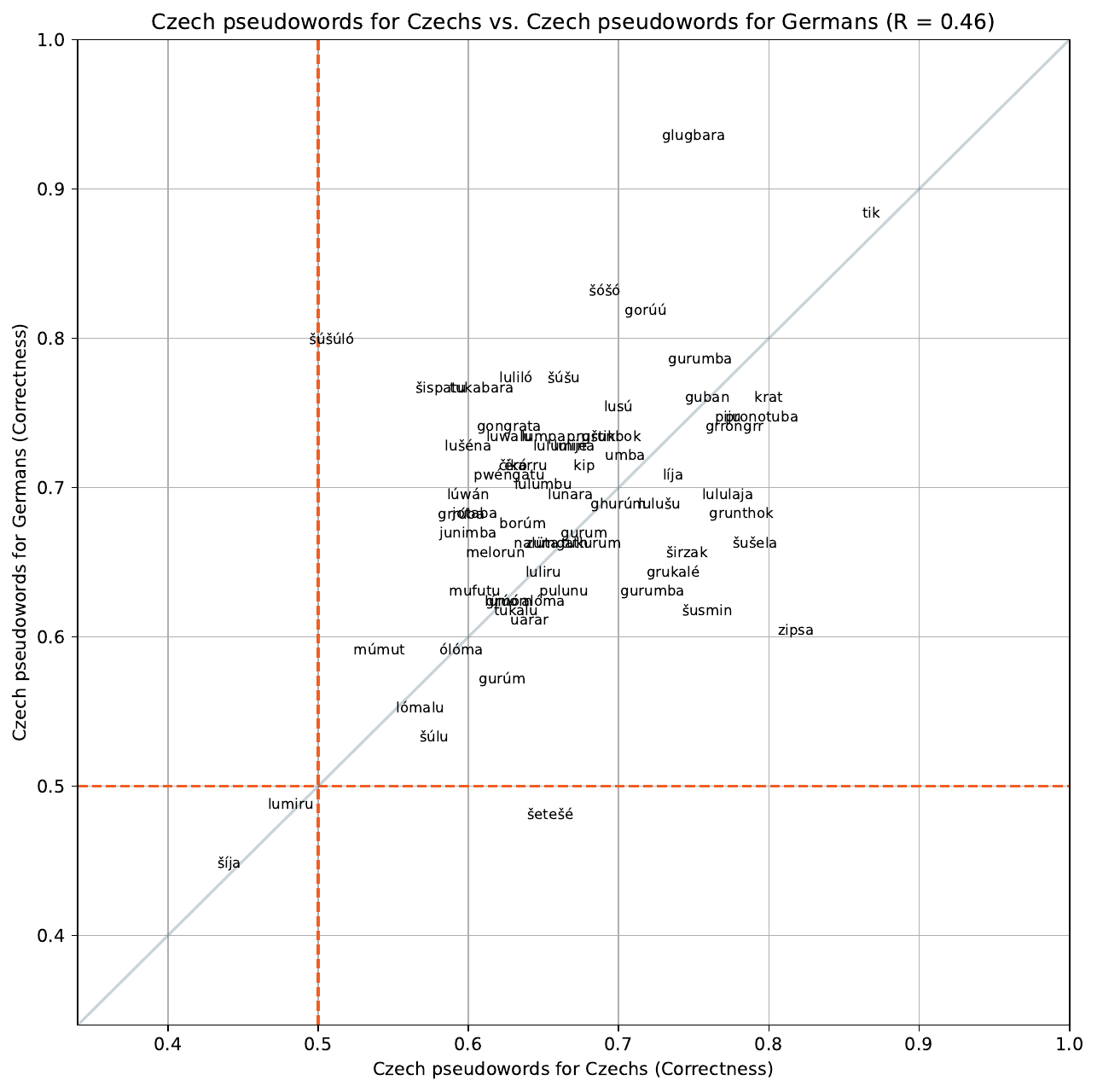}{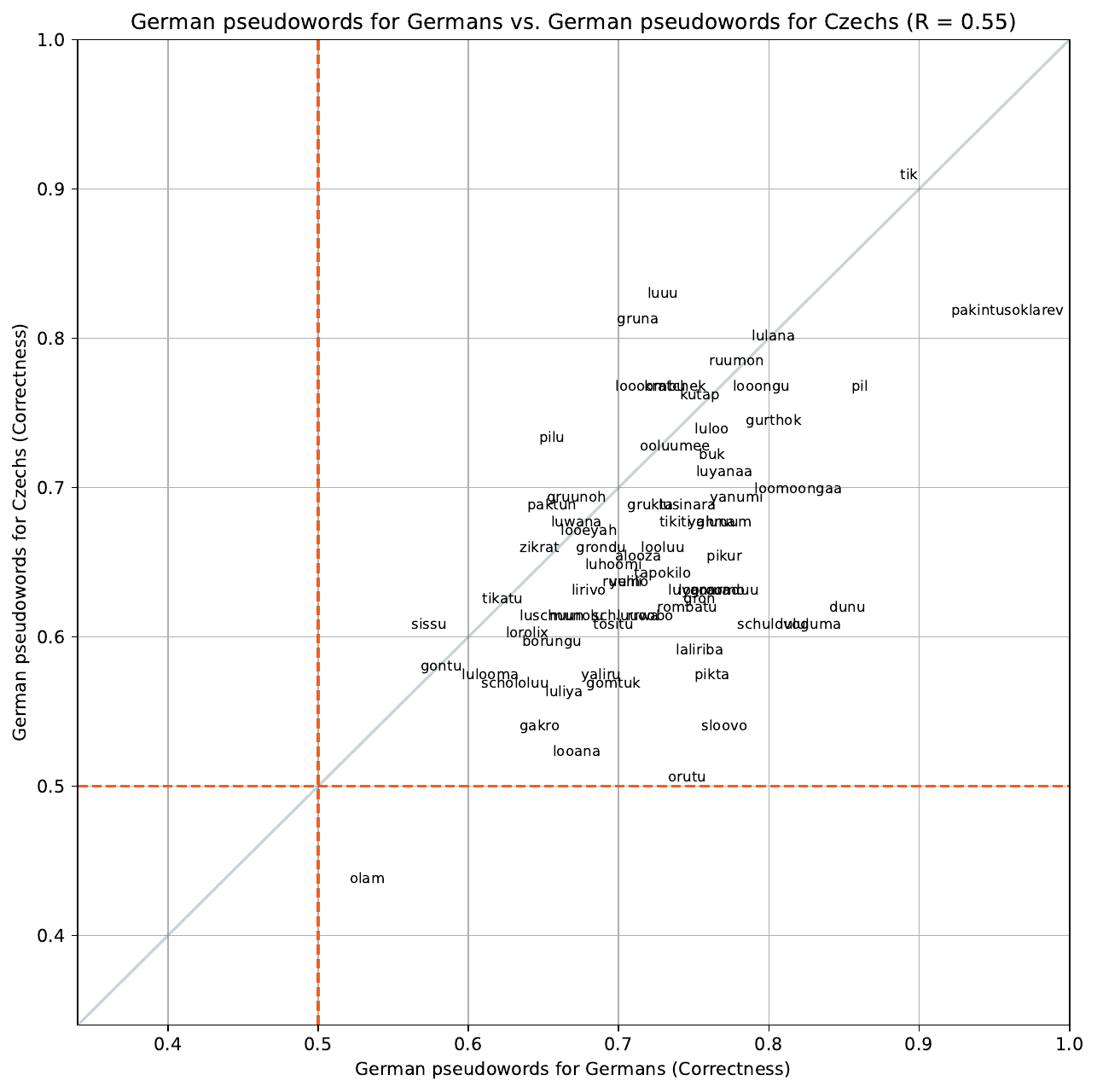}{ces-ger}{Czech prompted pseudowords presented to Czechs and Germans (top), and German prompted pseudowords presented to Czechs (bottom).}

\ChartComparison{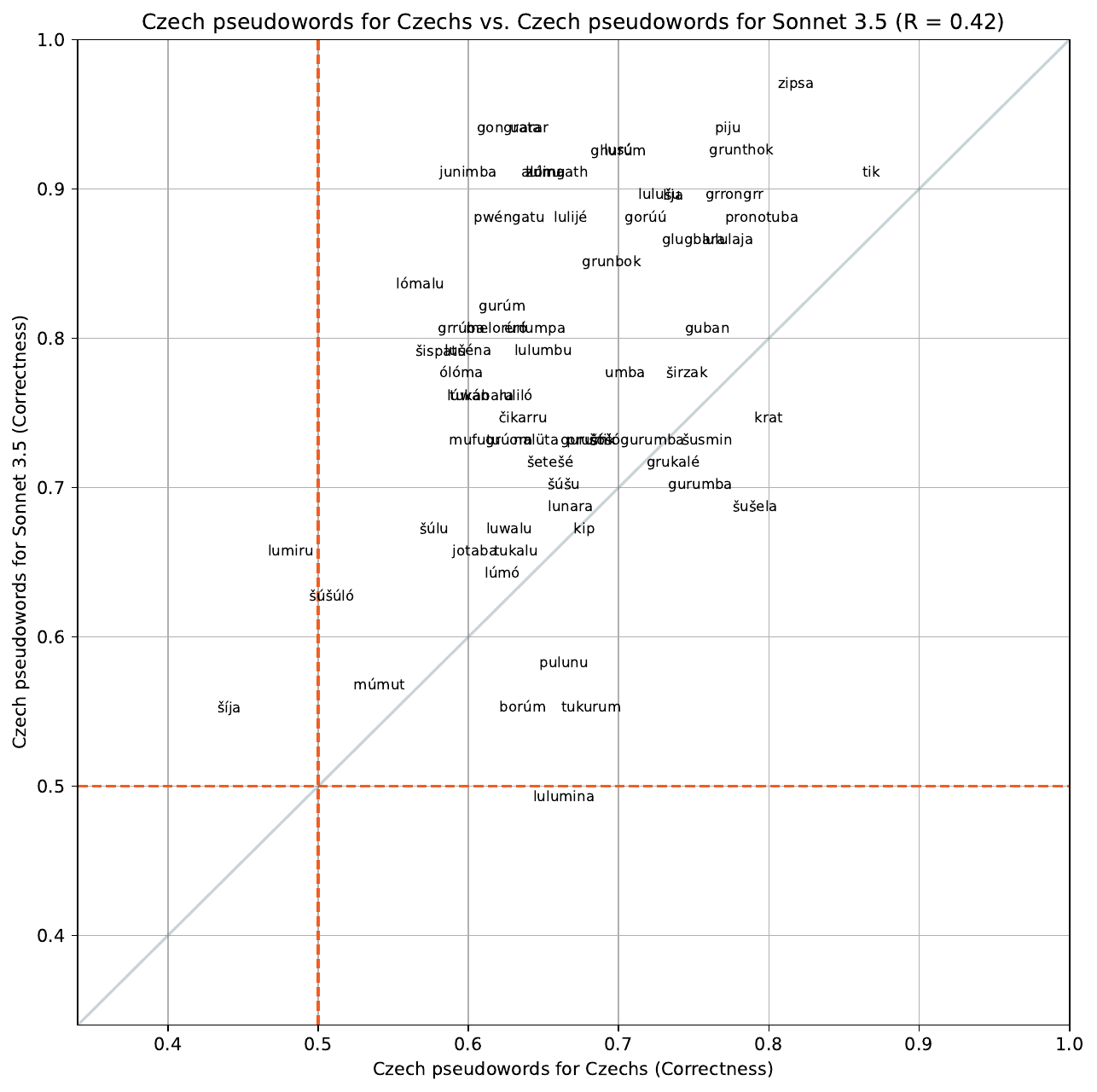}{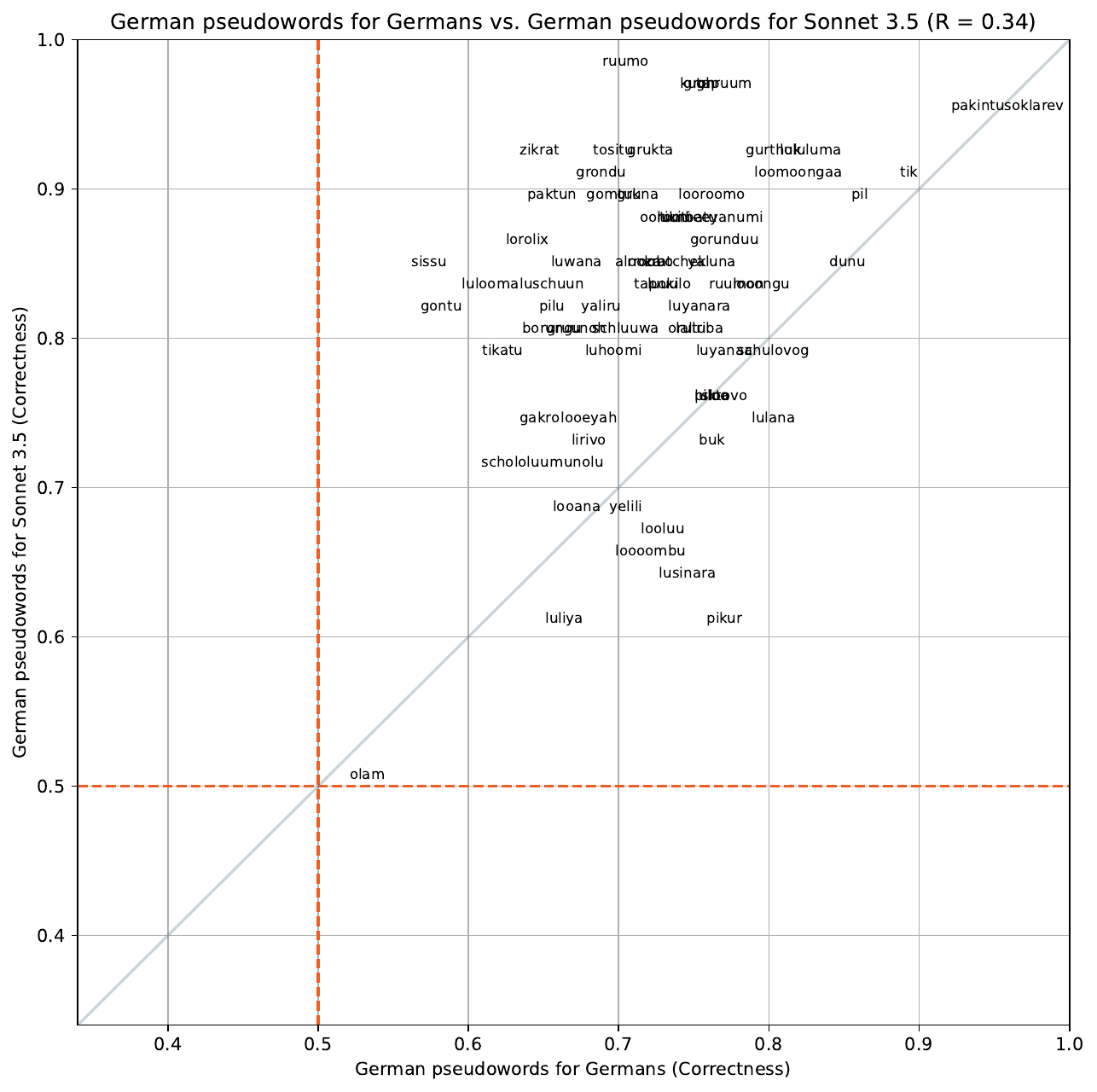}{hum-llm}{Czech prompted pseudowords presented to Czechs and Claude Sonnet 3.5 based participants (top), and German prompted pseudowords presented to Germans and Claude Sonnet 3.5 based participants (bottom).}

Similar differences and similarities can be observed between Czech and German participants, e.g., \textit{orutu} `direction' was quite easy for Germans but impenetrable for Czechs in the German dataset (Figure \ref{ChartComparison_Words_ces-ger}, bottom), while Czech prompted pseudoword \textit{šíja} `immaterial' was hard to guess for both Czechs and Germans (Figure \ref{ChartComparison_Words_ces-ger}, top).

Such patterns of agreement and disagreement are also found between human and LLM-based participants, e.g. German prompted pseudoword \textit{olam} `digestion' was hard to guess for Germans, Czech and also for Claude 3.5 Sonnet (Figure \ref{ChartComparison_Words_hum-llm}, bottom), but Czech prompted psoudoword \textit{lómalu} `ultimate good' was much easier for the LLM-based participants than for Czechs (Figure \ref{ChartComparison_Words_hum-llm}, top).

\begin{figure}[!tbp]
\centering
    \includegraphics[width=0.5\textwidth]{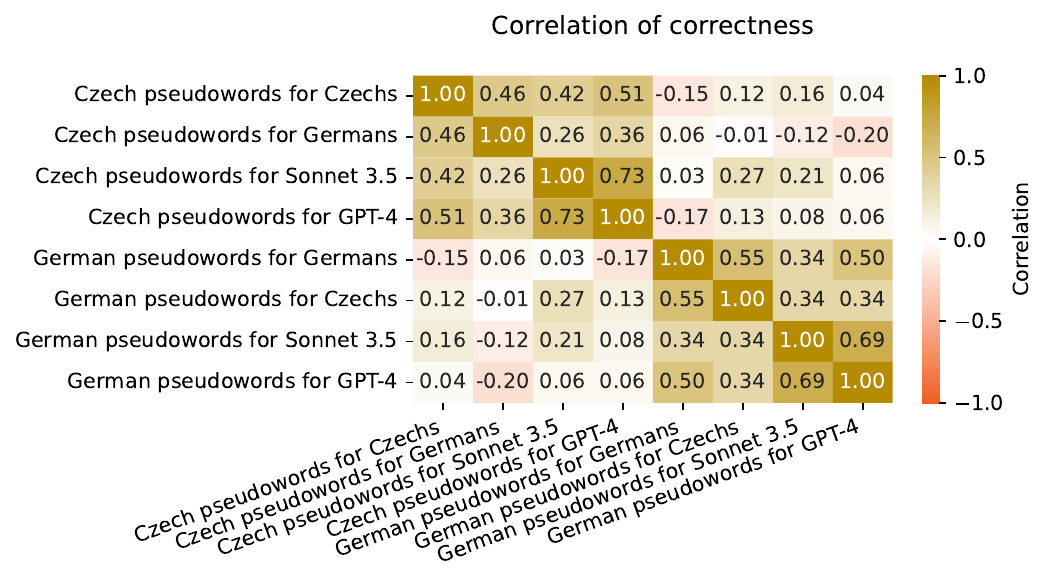}
\caption{Correlation matrix and heatmap across different groups of human and simulated participants.}
\label{Correlation}
\end{figure}

To quantify these differences and similarities, we employed Pearson's correlation coefficient, calculated for all relevant pairs and visualized in a heatmap (Figure \ref{Correlation}). The analysis reveals no correlation between the two artificial languages --- for instance, Czech-prompted pseudowords and German-prompted pseudowords both presented to German participants had a very low correlation of 0.06. However, when examining how different participant groups performed on the same pseudowords, correlations become notable or even quite high.

The strongest correlations are consistently found between LLM-based participants (73\% and 69\%), while lower correlations are between Czech and German human participants (46\% and 55\%). Human-LLM correlations show interesting variability: they can exceed human-human correlations (as in the case of Czech pseudowords with Czech and GPT-4 based participants at 51\%) or fall below them (German pseudowords shown to Germans and GPT-4 based participants at 50\%, which remains relatively high) or be quite low --- most other cases show correlations around 34\%, with the notable exception of Czech pseudowords shown to Germans and Claude 3.5 Sonnet, which exhibits the lowest correlation at 26\%.

The relatively high correlation between models and the overall higher correctness of LLM-based participants suggest that, while the models used clues for guessing that were comparable to those used by humans, they were also able to utilize certain clues that humans could not.

One of the main questions we asked was if LLMs will create language-specific or language-independent iconicity. We prompted the model in German, created what we called German-prompted pseudowords, and in Czech, creating Czech-prompted pseudowords. Our idea was that the models might take the language of the prompt as an inspiration and create the new words following patterns of the prompt language. This assumption was based also on our first attempts when prompting, when if we did not specify that the language is fictional and from a particular geographical area, the words were often basically Czech words (or German words, respectively) with additional syllable. To test this hypothesis, we had two groups of speakers, Czech and German, and half of each group guessed Czech-prompted, and half German-prompted pseudowords. In case that, e.g., Czech speakers would be significantly better in guessing Czech pseudowords than German pseudowords, and if they did so significantly better than the German speakers, we might assume that the iconicity models applied was language specific. However, our analysis did not show such effect. Czech-prompted pseudowords were guessed comparably well by Czech and by German speakers, and Czechs were not better at guessing Czech-prompted pseudowords than they were in guessing German-prompted ones. Similarly, German speakers were only slightly better in guessing German-prompted pseudowords than Czech-prompted pseudowords. Germans, however, were significantly better in guessing German-prompted pseudowords than Czech speakers. Nevertheless, there was no systematic pattern and Germans were overall better in guessing than Czechs for both datasets. 


\section{Conclusions}
The most remarkable finding of this study is that large language models encode not only arbitrary relations between symbols but also iconicity: despite both semantic and phonological qualities being only mediated and indirect for LLMs, they successfully connect them together.

LLM training extracts meta-linguistic information about iconicity and non-arbitrariness from their training data, making this information accessible both for generating highly iconic pseudowords and for guessing their meanings. When prompted to create iconicity-rich pseudowords in an artificial language, the LLM-based entity does not tailor the output according to the prompt language, but instead creates a language that is partially intelligible to human speakers of various languages.

The meanings of words in these LLM-generated artificial languages can be guessed not only by humans but also by LLM-based participants, even when based on a different model (Claude 3.5 Sonnet) than the one that generated the pseudowords (GPT-4). They are, in fact, much better than human participants in this task. Their results correlate with each other and also correlate with results of humans (slightly less, though).

The information about iconicity likely is not drawn solely from linguistic literature (which is present in the training data), as LLM-based entities do not justify their decisions by referencing such academic literature on iconicity, but rather explain their decisions similarly to humans (though this may be influenced by anthropomorphizing prompts). The studied models (GPT-4 and Claude 3.5 Sonnet) differ in how they justify their iconicity decisions. Their justifications when generating pseudowords typically differ from the justifications produced when guessing meanings, even within the same model (GPT-4).

The cues that humans use to guess meanings of unknown natural languages are also employed when guessing meanings in LLM-generated artificial languages, and these same cues are utilized by LLM-based participants (primarily length agreement and phonological similarity). However, there remains space for many other cues we could not explore (evidenced by the large intercept in the regression model) that may be specific to natural languages, human participants, or LLM-based participants.

Besides the theoretical implications, our findings also have practical applications. Typically, when using pseudowords in linguistic research, we work with made-up words that either follow or do not follow phonological rules of particular languages. However, these pseudowords are not created with particular meanings in mind; they are purely arbitrary sequences of characters. While existing tools for pseudoword generation offer word forms that sound more or less like natural language, they do not offer generation of words fitting suggested meanings. We can leverage this LLM ability to create experiments with pseudowords that can be more meaningfully compared to real words in a language, since real word forms often have some motivation in their meanings.




\section{Competing interests}
No competing interest is declared.

\section{Author Contributions Statement}
A.M. and J.M. planned the research, conceived and conducted the experiments, and wrote and reviewed the manuscript. J.M. also performed the statistical analyses and created the visualizations. L.R. programmed the online version of the experiment and collected the majority of data from German speakers.
L\makeL.L.B. and L.K. conducted the experiments, contributed to the research planning, and revised the manuscript.

\section{Funding}
Jiří Milička was supported by Czech Science Foundation Grant No. 24-11725S,
\url{gacr.cz}. 

Anna Marklová was supported by Primus Grant PRIMUS/25/SSH/010.

L\makeL udmila Lacková Bennet has been supported by the project JG\_2024\_020 implemented within the Palacký University Young Researcher Grant.

\section{Acknowledgments}
The authors thank all those who participated in collecting the data, namely 
Michaela Chodounská, 
Mia Miriam Levyová,
Kristína Harišová, and
Conor O'Neill.

\section{Declaration on using AI}
During the data analysis, the \emph{GPT-4}, \emph{GPT-4o}, and \emph{GPT-o1} models by \emph{OpenAI} and \emph{Claude 3.5 Sonnet} models by \emph{Anthropic} were consulted. However, all scripts underwent manual review and were, when necessary, corrected or further refined. By doing so, Jiří Milička assumes full responsibility for all the errors in these scripts. 

In the composition of this article, the \emph{GPT-4o}, \emph{GPT-o1}, \emph{Claude 3.5 Sonnet}, and \emph{Claude 3 Opus} were consulted for language styling and review. However, all ideas presented are original to the responsible authors, the generated text was further edited and underwent strict scrutiny.

\bibliographystyle{abbrvnat}
\bibliography{iconicity}


\end{document}